# Stochastic Optimization of Plain Convolutional Neural Networks with Simple methods

Yahia Saeed Assiri

Pace University, New York, USA

`ys34747w@pace.edu`

**Abstract.** Convolutional neural networks have been achieving the best possible accuracies in many visual pattern classification problems. However, due to the model capacity required to capture such representations, they are often oversensitive to overfitting and therefore require proper regularization to generalize well. In this paper, we present a combination of regularization techniques which work together to get better performance, we built plain CNNs, and then we used data augmentation, dropout and customized early stopping function, we tested and evaluated these techniques by applying models on five famous datasets, MNIST, CIFAR10, CIFAR100, SVHN, STL10, and we achieved three state-of-the-art- of (MNIST, SVHN, STL10) and very high-Accuracy on the other two datasets.

## 1    Introduction

Deep Learning has the main rule of the improvement in the field of computer vision, resulting in state-of-the-art performance in many challenging tasks [4], such as object recognition [2], semantic segmentation [1], image captioning [11], and human pose estimation [12]. Using the convolutional neural networks (CNNs) [10] was the main reason for many of the significant accomplishments over the past few years because they can learn deep feature representations of images. However, as complexity increases, the resource utilization of such models also increases. Modern deep networks contain hundreds of millions of parameters which provide the necessary representational power for such tasks [3], as a result of the huge representational power the probability of overfitting increases and leads to poor generalization. To fight the overfitting, different regularization techniques can be applied, such as data augmentation, dropout, and early stopping. Data augmentation is a very famous technique due to its ease of implementation and effectiveness. Using image transforms such as rotation shearing scaling cropping, or flipping can be applied to create new data which can be used to improve accuracy [10]. Large models can also be regularized by adding noise during the training process, whether it be added to the input [4], weights, or gradients. One of the most common uses of noise for improving model accuracy is dropout [8], which stochastically drops neuron activations during training and, as a result, discourages the co-adaptation of feature detectors. In this work, we consider applying different data augmentation techniques combined with dropout. These techniques encourage plain convolutional networks to achieve better generalization and get better results in the validation and testing phase.

In the remainder of this paper, we introduce collected optimization methods and demonstrate that using data augmentation and dropout can improve model robustness and lead to better model performance. We show that these simple methods work with a simple plain convolutional neural networks model and can also be combined with most regularization techniques, including learning rate and early stopping and other regularization techniques in a very simple manner.

## 2 Data Augmentation for Image

Data augmentation is an effective technique for improving the accuracy of modern image classifiers, and it has long been used in practice when training convolutional neural networks. LeCun et al. used many affine transformations, like horizontal and vertical translation, scaling, squeezing, and shearing, to improve their model's accuracy when training LeNet5 [10].

Krizhevsky did apply image mirroring, cropping, as well as randomly adjusting color and intensity values based on ranges determined using principal component analysis on ImageNet dataset to improve the performance of AlexNet [8] for the 2012 ImageNet Large Scale Visual Recognition Competition [4].

Bengio et al. [5] did apply a large variety of transformations to a handwritten character dataset, for example, local elastic deformation, motion blur, Gaussian smoothing, Gaussian noise, salt and pepper noise, pixel permutation and affine transformations[4].

When training Deep Image [13] on the ImageNet dataset, Wu et al. did apply a wide range of color casting, vignetting, rotation, and lens distortion, as well as horizontal and vertical stretching.

Lemley et al.came up with a learned end-to-end approach called Smart Augmentation [6] instead of relying on hard-coded transformations. In this method, a neural network is trained to intelligently combine existing samples to generate additional data that is useful for the training process. [4].

DeVries et al. came up with a new technique called Cutout, which is [4] closest to the occlusions technique. However, occlusions generally take the form of scratches, dots, or scribbles that overlay the target character, while cutout use zero-masking to completely obstruct an entire region. Cutout can be interpreted as applying a spatial dropout in input space, much in the same way that convolutional neural networks leverage information about spatial structure to improve performance over that of feedforward networks.

# 3 Dropout in Neural Networks

The second common regularization technique that we used in our models is dropout [8], which was first introduced by Hinton et al., they used the method with a range of different neural networks on different problem types achieving improved results, including handwritten digit recognition (MNIST), photo classification (CIFAR-10), and speech recognition (TIMIT)[20]. Dropout is implemented by setting hidden unit activations to zero with some fixed probability during training. All activations are kept when evaluating the network, but the resulting output is scaled according to the dropout probability. This technique has the effect of approximately averaging over an exponential number of smaller sub-networks and works well as a robust type of bagging, which discourages the co-adaptation of feature detectors within the network. [4]

Nitish Srivastava et al. [16] used dropout on a wide range of computer vision, speech recognition, and text classification tasks and found that it consistently improved performance on each problem.

George Dahl et al. [19] used a deep neural network with rectified linear activation functions and dropout to achieve state-of-the-art results on a standard speech recognition task [20].

Tompson et al. introduce Spatial Dropout [14], which randomly discards entire feature maps rather than individual pixels.

While dropout was found to be very effective at regularizing fully connected layers, we discovered that it does have the same powerful at convolutional neural when used at the best place and when used with the best rate.

# 4 Early Stopping

Early stopping is a form of regularization used to avoid overfitting when using some methods, such as gradient descent. Such methods update the learner to make it better fit the training data with each iteration. The early stopping rule is to stop the model training at a certain number of iterations to keep the validation dataset free of overfitting. Early stopping rules have been employed in many different machine learning methods, with varying amounts of theoretical foundation.

# 5 Implementation Details

## 5.1 Architecture Concept

One consideration here is the number of weights that can be set independently, the more of those we have, the greater the risk of overfitting, and the greater the training time. So, increasing this number makes training take longer and runs a higher risk of overfitting, but potentially increases the expressiveness of our neural network. We did want the number to be as small as possible without sacrificing accuracy. So, trying something small and increase it until the model stops getting improving in the accuracy, we also considered this concept when setting the dropout rate.

### 5.2    Architecture and Design

As a result of this concept, We built simple convolutional network models[9] (plain convolutional neural networks with no residual blocks ), four layers for MNIST dataset, and 11 layers for Cifar10, Cifar100, SVHN datasets and 13 layers for STL10, The networks employ a homogeneous design utilizing 3×3 kernels for convolutional layer and 2×2 kernels for max-pooling operations, the number of model layers depends on the size of input image, for instance, MNIST has 28*28 grayscale images, this reduced the number of layers to only four Layers; however, STL10 has 96*96*3 in this case the number of layers is 13.  Applying max-pooling (2*2) after every two convolutional layers helped the model not only to control the depth of the model but also reduced variance and reduced computation complexity (as 2*2 max-pooling/average pooling reduce 75% of data), and extract low-level features from neighborhood, Relu activation function was added before every max-pooling layer, and there is a dropout layer following every max-pooling and then end the model with wide classifier layer which is composed of two fully connected layers followed by dropout layer and then the last layer which is the softmax layer.

The learning rate is 0.01 and the batch sizes used are different upon the dataset's image size, for the MNIST dataset, we found that large batch size is very useful and reduced the fluctuating on the validation set, so the batch size for MNIST dataset was 256 and for Cifar10 Cifar100 and SVHN batch size was 128, however for STL10 the batch size was only 8, we considered the size of the image (96*96), so we used a trade-off strategy, if the image size is very big the batch size is small however when the image size is small, the batch size becomes larger.

### 5.3    Our customized early stopping

We built a naive early stopping to watch the validation curves, we built it in order to control the fluctuating phenomenon by making a baseline point and then asking the model not to stop unless passing this point, and that helped the models to catch highest possible accuracy

### 5.4    Augmentation techniques

After applying the trial and error method, we combined five data augmentation techniques, Table 6, and we found that flipping techniques did not help the model to generalize during training, they negatively affected the model; as a result, we did not use them.

## 6    Experiments on MNIST

MNIST dataset Lecun et al. (1998) [7] consists of 70,000 28x28 grayscale images of handwritten digits 0 to 9, of which 60,000 are used for training, and 10,000 are used for testing.

After we formulated our model, we conducted our experiments on MNIST through two stages, The first stage is applying three paradigms of Dropout technique, Table 1, in the first paradigm we applied regular dropout after the fully connected layer on our model without inserting any dropout layer in the convolutional part, second paradigm we applied spatial Dropout before each max-pooling layer in the part of the convolutional network without applying dropout after the fully connected layer, and finally, in the third paradigm, we applied mixed technique which is a combination of inserting Spatial Dropout before each max-pooling layer and also inserting regular dropout after the fully connected layer, and the number of the epoch is 2500, Fig 1, we achieved very high accuracy on all three paradigms, Table 1, however, the highest accuracy was achieved is by applying the Dropout only after the fully connected layer.

**Table 1.** classification error of 3 methods applied to Dropout on MNIST.

| method | Dropout ratio | Accurecy% | Error rate% |
|---|---|---|---|
| Only regular dropout After FC | p=0.4 | 99.79 | 0.21 |
| Only Spatial Dropout before each maxpooling layer | p=0.125 | 99.78 | 0.22 |
| Combined dropout(Spatial Dropout before each maxpooling+ regular dropout After FC) | ps=0.125 pr=0.4 | 99.76 | 0.24 |

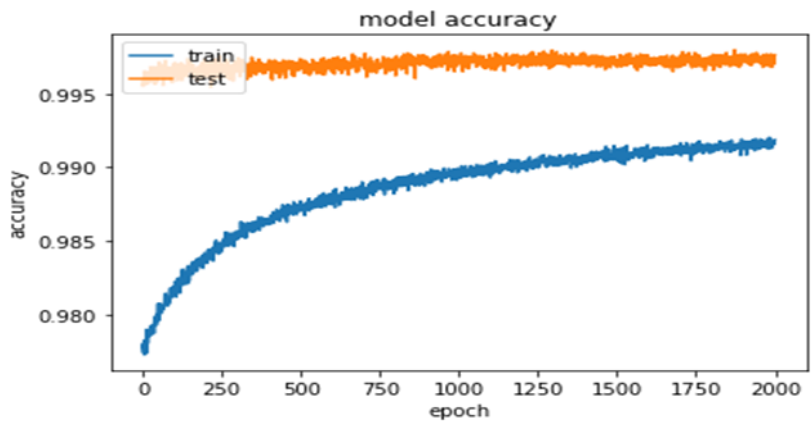

**Fig. 1.** MNIST Accuracy –first 2000 epoch before the early stopping condition achieved (all three paradigms needed as minimum 2000 epoch).

To test the robustness of Dropout we considered the results that we achieved through the first stage and chose the paradigm which only has the dropout layer after the fully connected layer, and then started the second stage which aimed to figure out what is the best ratio of regular dropout that we should use and what is the size of fully connected layer considered the best to be selected.

We found there is a direct proportion of the size of the fully connected layer and the ratio of dropout out, so we conducted this stage by gradually and simultaneously increasing the size of the fully connected layer and the rate of dropout.

We found the best size of the fully connected layer is 2048, and the best ratio of the dropout is 0.8, and to make sure of the experiment´s outcome, we tested this algorithm five times with the same parameters as shown in Table 2.

Our experiments achieved a new state of the art, Table 3, of all five runs, the highest accuracy achieved was 99.83 % with error rate only 0.17 %.

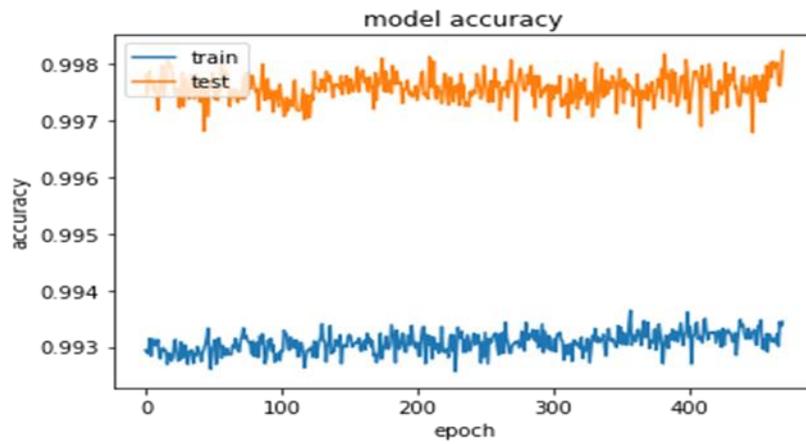

**Fig. 2.** MNIST Accuracy-Where the customized early stopping happened for the regular Dropout experiment, the model strongly generalized with FC =2048 and dropout=0.8.

**Table 2.** results of all five experiments done on MNIST by using FC=2048 and Dropout= 0.8.

| Experiments # | Dropout rate=(0.8) | Size of the Fully connected Layer | % Accurecy | % error |
|---|---|---|---|---|
| 1 | 0.8 | 2048 | 99.81 | 0.19 |
| 2 | 0.8 | 2048 | 99.82 | 0.18 |
| 3 | 0.8 | 2048 | 99.82 | 0.18 |
| 4 | 0.8 | 2048 | 99.82 | 0.18 |
| **5** | **0.8** | **2048** | **99.83** | **0.17** |

**Table 3.** Top MNIST Results.

| Method | Error rate |
|---|---|
| DropConnectWan et al. (2013) [7] | 0.21% |
| Multi-column DNN for Image Classification Ciregan et al. (2012) [18] | 0.23% |
| APAC Sato et al. (2015) [17] | 0.23% |
| Generalizing Pooling Functions in CNN Lee et al. (2016) [15] | 0.29% |
| Stochastic Optimization of Plain Convolutional Neural Networks with simple methods | 0.17% |

## 7    Experiments on other Data Sets

We used four datasets, in addition to MNIST, to evaluate our models SVHN, CIFAR-10, CIFAR-100, and STL10. These data sets include different image types and training set sizes. Our Models achieved state-of-the-art results on three datasets MNIST, SVHN, and STL10, Table 4.

### 7.1    CIFAR10 AND CIFAR100

Both of the CIFAR datasets consist of 60,000 color images of size 32×32 pixels. CIFAR-10 has ten distinct classes, such as cat, dog, car, and boat. CIFAR-100 contains 100 classes [4].

For those two datasets, we applied dropout to convolutional neural networks. The best architecture that we found has 11 convolutional layers, followed by two fully connected layers. All activation functions used are ReLUs. A max-pooling layer followed each convolutional layer, and dropout was applied after the max-pooling to all layers of the network with the probability of (0.25).

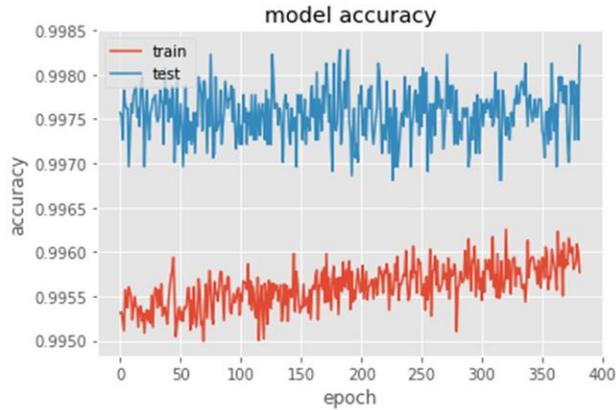

**Fig. 3.** MNIST Accuracy-Where the customized early stopping happened for the 5th experiment (the error rate on the test set =0.17%).

### 7.2 SVHN

The Street View House Numbers (SVHN) dataset contains a total of 630,420 color images with a resolution of 32×32 pixels. Each image is centered on a number from one to ten, which needs to be identified. The official dataset split contains 73,257 training images and 26,032 test images, but there are also 531,131 additional training images available, we used both available training sets when training our models. For this data set, we applied regular dropout to the convolutional neural networks part, and we did not use the spatial dropout in this experiment. The best architecture that we found has 11 convolutional layers, followed by two fully connected layers. All activation units were ReLUs. A max-pooling layer followed each convolutional layer, and dropout was applied to all the layers of the network with the probability of (0.25); the fully connected layer size was 1024, followed by a dropout with ratio 0.4.

We achieved new state the art for SVHN if only using Plain Convolutional Neural Networks, with no residual block, we achieved 98.50 % with an only error rate of 1.5%, and when comparing with residual neural networks, our networks achieved the fourth place.

**Table 4.** Total experiments result in all five datasets.

| Dataset | %Accu-racy | %error | The total of model params |
|---------|-----------|--------|---------------------------|
| MNIST | 99.83 | 0.17 | >1,400.000 |
| Cifar10 | 94.29 | 5.71 | 4,252,298 |
| Cifar100 | 72.96 | 27.04 | >4,252,298 |
| SVHN | 98.50 | 1.50 | 4,252,298 |
| STL10 | 88.08 | 11.92 | >5,000,000 |

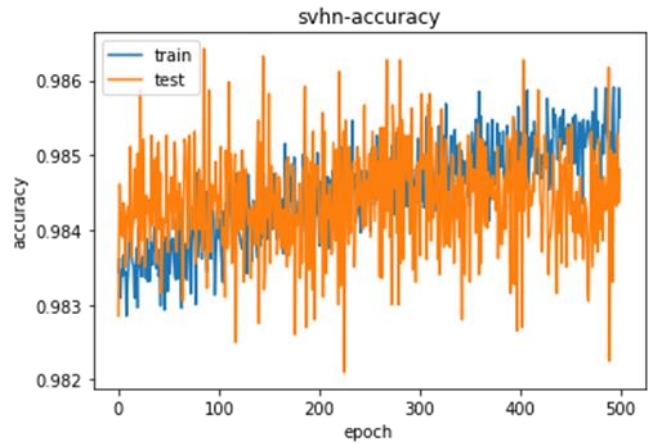

**Fig. 4.** SVHN Accuracy.

### 7.3 STL-10

The STL-10 dataset [4] consists of a total of 113,000 color images with a resolution of 96×96 pixels. The training set only contains 5,000 images, while the test set consists of 8,000 images. All training and test set images belong to one of ten classes, such as airplane, bird, or horse. The remainder of the dataset is composed of 100,000 unlabeled images belonging to the target ten classes, plus additional but visually similar classes. While the main purpose of the STL-10 dataset is to test semi-supervised learning algorithms, we used it to observe how our collected optimization method performs when applied to higher resolution images with a small training set. For this reason, we did not use the unlabeled portion, which is the larger portion of the dataset, and only used the labeled training set.

For this data set, we applied regular dropout to the convolutional neural networks part, and we did not use the spatial dropout in this experiment. The best architecture that we found has 13 convolutional layers, followed by two fully connected layers. All activation units were ReLUs. A max-pooling layer followed each convolutional layer, and dropout was applied to all the layers of the network with the probability of (0.25), the fully connected layer size was 1024, followed by a dropout with ratio 0.4. Our model was able to achieve the best accuracy ever on supervised classification of STL-10 and was able to defeat cut out [4] without "cutout" tech, and without "residual block," the model achieved 88.08% with only error rate 11.92%, Table 5.

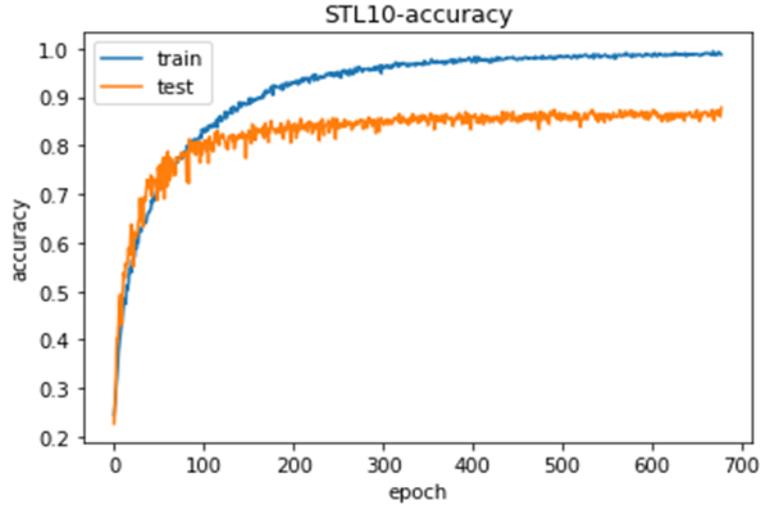

**Fig. 5.** STL10 Accuracy.

**Table 5.** Datasets of the new State of the art achieved.

| Dataset | Accuracy% | % error | New state of the art? | The previous state of the art (%error) |
|---------|-----------|---------|-----------------------|----------------------------------------|
| MNIST | 99.83 | 0.17% | Yes | DropConnect Wan et al.[7] (2013) (0.21%) |
| STL10 | 88.08 | 11.92 % | Yes | Improved Regularization with cutout Cutout. (2017)[4] (12.74±0.23)% |
| SVHN | 98.50 | 1.50% | Yes (in plain Cnn) Fourth overall | Generalizing Pooling Functions in Convolutional Neural Networks: Mixed, Gated, and Tree(2016)[15] (1.69%) |

**Table 6.** Augmentation techniques used.

| technique | Use? |
|---|---|
| rotation | Only used with mnist |
| shearing | Yes |
| Shifting up and down | Yes |
| Zooming | Yes |
| rescale | Yes |
| cutout | No |
| flipping | No |

## 8    Conclusion

The Simple Plain Convolutional Neural Networks model is still able to achieve outstanding performance and to compete for the residual Neural networks technique,and when comparing the number of trainable parameters of both techniques we can recognize how valuable using the plain model, the parameters numbers reduced by %80 and CNNs, the smaller simple models, able to get the best performance for some datasets.